\documentclass[10pt]{article} 

\usepackage[preprint]{rlj} 

\usepackage{amssymb}            
\usepackage{mathtools}          
\usepackage{mathrsfs}           
\usepackage{graphicx}           
\usepackage{subcaption}         
\usepackage[space]{grffile}     
\usepackage{url}                
\usepackage{lipsum}             
\usepackage[]{algorithm2e}
\usepackage{tikz}
\usetikzlibrary{automata, positioning}
\usetikzlibrary{decorations.text, decorations.shapes}
\usetikzlibrary{decorations.pathmorphing}

\usepackage{tabularx}
\usepackage{array}
\newcolumntype{Y}{>{\raggedright\arraybackslash}X}
\usepackage{subcaption}
\usepackage{multirow}
\usepackage{booktabs}       
\usepackage{float}

\title{Mixture of Autoencoder Experts Guidance using Unlabeled and Incomplete Data for Exploration in Reinforcement Learning}

\setrunningtitle{Mixture of Experts Guidance using Unlabeled and Incomplete Data for Exploration}

\author{Elias Malomgré\textsuperscript{1},\ \  Pieter Simoens \textsuperscript{1}}

\emails{elias.malomgre@ugent.be, \ \ pieter.simoens@ugent.be}

\affiliations{
$^{1}$\textbf{IDLab, Ghent University-imec, Technologiepark-Zwijnaarde 126, 9052 Ghent, Belgium}
}

\contribution{
    This paper introduces MoE-GUIDE, an RL framework that learns from incomplete, unlabeled, and imperfect expert demonstrations by using a Mixture of Autoencoders as a similarity model.
}{
    Prior work on leveraging demonstrations in RL typically assumes complete and high-quality demonstrations, making them less applicable to realistic scenarios with partial or noisy data.
}

\contribution{
    We propose a mapping function that transforms state similarity with expert data into a shaped intrinsic reward, enabling flexible and targeted exploration.
}{
    Existing intrinsic motivation approaches often rely on unprocessed intrinsic rewards, which can make exploration hard to control or insufficiently focused.
}

\contribution{
    We demonstrate that MoE-GUIDE enables robust exploration and strong performance in both sparse and dense reward environments, even when expert data is limited, incomplete, or partially observed.
}{
    Prior methods tend to degrade in performance when provided with sparse or imperfect demonstrations, whereas our method maintains effectiveness across a range of challenging settings.
}


\summary{Recent trends in Reinforcement Learning (RL) highlight the need for agents to learn from reward-free interactions and alternative supervision signals, such as unlabeled or incomplete demonstrations, rather than relying solely on explicit reward maximization. Developing generalist agents that can adapt efficiently in real-world environments often requires leveraging these reward-free signals to guide learning and behavior. While intrinsic motivation techniques provide a means for agents to seek out novel or uncertain states in the absence of explicit rewards, they are often challenged by dense reward environments or the complexity of high-dimensional state and action spaces. Furthermore, most existing approaches rely directly on the unprocessed intrinsic reward signals, which can make it difficult to shape or control the agent’s exploration effectively. We propose an approach that can effectively utilize expert demonstrations, even when they are incomplete and imperfect. By applying a mapping function to transform the similarity between an agent’s state and expert data into a shaped intrinsic reward, our method allows for flexible and targeted exploration of expert-like behaviors. We employ a Mixture of Autoencoder Experts to capture a diverse range of behaviors and accommodate missing information in demonstrations. Experiments show our approach enables robust exploration and strong performance in both sparse and dense reward environments, even when demonstrations are sparse or incomplete. This provides a practical framework for RL in realistic settings where optimal data is unavailable and precise reward control is needed.
}

\begin{document}

\maketitle  

\begin{abstract}
Recent trends in Reinforcement Learning (RL) highlight the need for agents to learn from reward-free interactions and alternative supervision signals, such as unlabeled or incomplete demonstrations, rather than relying solely on explicit reward maximization. Additionally, developing generalist agents that can adapt efficiently in real-world environments often requires leveraging these reward-free signals to guide learning and behavior. However, while intrinsic motivation techniques provide a means for agents to seek out novel or uncertain states in the absence of explicit rewards, they are often challenged by dense reward environments or the complexity of high-dimensional state and action spaces. Furthermore, most existing approaches rely directly on the unprocessed intrinsic reward signals, which can make it difficult to shape or control the agent’s exploration effectively. We propose a framework that can effectively utilize expert demonstrations, even when they are incomplete and imperfect. By applying a mapping function to transform the similarity between an agent’s state and expert data into a shaped intrinsic reward, our method allows for flexible and targeted exploration of expert-like behaviors. We employ a Mixture of Autoencoder Experts to capture a diverse range of behaviors and accommodate missing information in demonstrations. Experiments show our approach enables robust exploration and strong performance in both sparse and dense reward environments, even when demonstrations are sparse or incomplete. This provides a practical framework for RL in realistic settings where optimal data is unavailable and precise reward control is needed.
\end{abstract}

\section{Introduction}
The pursuit of intelligent, adaptive agents in reinforcement learning (RL) increasingly requires methods that go beyond traditional reward maximization. In many real-world settings, agents must learn and generalize from limited or ambiguous feedback, such as sparse environmental rewards, incomplete demonstrations, or unlabeled experience. These scenarios highlight the need for RL approaches that can leverage alternative signals, whether from the environment or from human guidance, to form robust representations and discover useful behaviors.

A key strategy for enabling learning in such settings has been the use of intrinsic motivation, which encourages agents to seek out novel, uncertain, or otherwise informative states. Techniques such as curiosity-driven exploration \cite{ICM} and Random Network Distillation (RND) \cite{RND} provide intrinsic rewards based on prediction errors or novelty estimates, guiding agents through unfamiliar regions of the state space. More recently, autoencoder-based methods have emerged as powerful tools for quantifying state familiarity, using reconstruction loss to identify and reward visits to underexplored or novel states \cite{klissarov2019variational, kubovvcik2023signal, yan2024autoencoder, liu2019state}. While these approaches have demonstrated effectiveness in sparse-reward environments, they depend critically on learning meaningful representations, which can be challenging in high-dimensional, continuous environments \cite{aubret2019survey}.

Demonstrations can hold valuable information, and therefore, approaches like Behavior Cloning \cite{BC} directly imitate expert trajectories but lack the flexibility to improve beyond suboptimal data. Inverse reinforcement learning (IRL) \cite{IRL} and guided exploration methods \cite{Gail} infer reward functions or policies from demonstrations, allowing for some autonomy; however, they most of the time do not utilize extrinsic rewards and struggle to outperform the expert. Demonstrations are also used in RL to improve performance by adding them to the replay buffer \cite{paine2019making, rajeswaran2017learning} or by leveraging BC to jump-start the policy or as a guide during the training process \cite{paine2019making, rajeswaran2017learning}. Importantly, these approaches typically assume access to complete trajectory data, including actions and next states. In practice, however, obtaining such complete datasets is challenging. Technical limitations, privacy concerns, and sensor issues frequently result in missing, incomplete, or noisy demonstration data \cite{rao2018origin, zhao2019distributionally, cao2023jointly}. This is common in domains like robotics and traffic modeling, where actions or states may only be partially observed or where data is sparse due to sensor failures or limited coverage \cite{BCO, wei2020learning, sun2019adversarial, xu2021arail}. Collecting high-quality, dense expert data is often costly, time-consuming, or impossible, and may still yield imperfect demonstrations \cite{camacho2021sparsedice}. Demonstrations may also be sparse or imperfect, such as those from low-bitrate videos. Consequently, methods that require fully observable, dense demonstrations may perform poorly in real applications. This highlights the need for research on methods that can learn effectively from incomplete, state-only, or imperfect demonstrations. However, since we do not have access to complete trajectories and can only observe some states that the expert visited, we cannot provide the agent with explicit expert paths. Therefore, the agent must infer the expert's path by navigating between the states likely visited by the expert, which is a much more challenging but realistic scenario.

Although state-only IL and IRL methods can be used to enrich extrinsic rewards with incomplete and unlabeled data, guiding the agent to regions in the observation space, these reward models will be computationally expensive to obtain, as they require numerous interactions with the environment \cite{10602544}. Therefore, our method utilizes state-only expert demonstrations that can have gaps to train a model, such as an autoencoder, density estimator, or RND, that measures the similarity between current states and expert behavior, thereby producing a loss landscape over the observation space. We introduce a mapping function that normalizes model loss into an intrinsic reward, ranging from 0 to 1. States with losses below the minimum threshold are considered expert-like and receive the highest reward. States with losses above the maximum threshold receive zero reward. The reward is calculated using a chosen mapping function, such as linear or exponential, for losses in between. This approach enables precise control over the reward structure, allowing us to eliminate undesirable local minima and encourage exploration in regions likely to yield expert-like outcomes.

An essential aspect of this framework is the similarity model’s ability to distinguish expert behavior from random states. Our research found that standard autoencoders with narrow bottlenecks can be highly selective for expert data, as they focus solely on extracting useful features to optimize the reconstruction of expert behavior. Therefore, we focused our research on this model. However, as in the case of autoencoders, a single similarity model may still struggle to capture the full diversity of expert demonstrations. To address this, we introduce Mixture of Experts Guidance using Unlabeled and Incomplete Data for Exploration (MoE-GUIDE), a mixture-of-experts model using several similarity models, each specializing in different features or modes of the expert data. A gating network combines its outputs, dynamically weighting each expert for a given state. This forms well-defined regions in the observation space, similar to expert-like states. By converting this landscape into an intrinsic reward, the agent is guided toward regions aligned with expert experience, thereby improving exploration efficiency.

\section{Background}

\subsection{Reinforcement Learning Beyond Rewards}

Traditional Reinforcement Learning (RL) is grounded in the Markov Decision Process (MDP) framework, where an agent interacts with an environment by observing states $s \in \mathcal{S}$, selecting actions $a \in \mathcal{A}$, and receiving rewards $r \in \mathbb{R}$ defined by a reward function $R: \mathcal{S} \times \mathcal{A} \to \mathbb{R}$ \cite{sutton1998reinforcement}. The agent's objective is typically to maximize the expected cumulative discounted return:
\begin{equation}
    \mathbb{E}_\pi\left[\sum_{t=0}^{\infty} \gamma^t r(s_t, a_t)\right],
\end{equation}
where $\gamma \in [0, 1)$ is the discount factor.

However, real-world environments often lack well-specified, dense, or even meaningful reward signals. This has motivated a growing body of research on reward-free RL \cite{jin2020reward}, where agents learn from alternative forms of supervision, such as unlabeled interaction data, expert demonstrations, preferences, or implicit human feedback. Reward-free RL aims to develop agents that can acquire generalizable skills and representations from environmental structure or diverse signals, thereby facilitating rapid adaptation when task rewards become available or when rewards are difficult to specify.

\subsection{Representation Learning and Intrinsic Motivation}

A core challenge in reward-free RL is learning meaningful representations and skills from unlabeled data. Intrinsic motivation offers one solution, providing internal reward signals that incentivize exploration, skill development, or the acquisition of predictive representations.

\textbf{Prediction- and surprise-based methods} reward the agent for visiting novel or unpredictable states. The Intrinsic Curiosity Module (ICM) \cite{ICM} measures the prediction error of a learned forward model as an intrinsic reward, thereby encouraging the agent to seek out transitions that are difficult to predict. Similarly, Random Network Distillation (RND) \cite{RND, RND2} assesses state novelty by comparing the output of a fixed random network to that of a predictor network, guiding exploration toward poorly represented states.

\textbf{Novelty- and count-based strategies} encourage agents to visit rarely encountered states, either through explicit state visitation counts in discrete domains or via pseudo-counts and density models in high-dimensional spaces \cite{bellemare2016unifying, ostrovski2017count, zhao2019curiosity}. These approaches increase the diversity of experiences and can improve sample efficiency in sparse-reward environments.

\subsection{Learning from Demonstrations and Alternative Signals}

When reward functions are ill-defined or unavailable, demonstrations and other human-centric signals can serve as crucial supervisory information. Demonstration-driven techniques, such as imitation learning \cite{Gail}, inverse reinforcement learning \cite{IRL}, and learning from observation \cite{gailfo, zhu2020off}, leverage expert trajectories or behavioral cues to shape agent behavior. These methods can guide the learning process of an agent.

Recent advances have addressed challenges such as incomplete, suboptimal, or action-free demonstrations \cite{wei2020learning, xu2021arail, camacho2021sparsedice, fu2017learning}. However, these methods do not account for extrinsic rewards, and therefore, demonstrations have also been integrated into off-policy RL via replay buffers \cite{paine2019making, rajeswaran2017learning}. Other methods use hand-crafted reward terms based on demonstrations \cite{peng2018deepmimic} and leverage BC to pretrain the policy or to guide the learning process \cite{nair2018overcoming, rajeswaran2017learning}. However, these methods assume complete data that exhibits near-perfect behavior. Therefore, our method proposes a framework to guide the learning process of an agent with incomplete, unlabeled, and imperfect demonstrations. 

\subsection{Soft Actor-Critic}

Our method builds upon the Soft Actor-Critic (SAC) framework \cite{SAC1, SAC2}, an off-policy actor-critic algorithm that augments the reward maximization objective with an entropy maximization term. This encourages diverse behavior and robust exploration:
\begin{equation}
J(\pi) = \mathbb{E}_{\tau \sim \rho_\pi}\left[\sum_{t=0}^T r(s_t, a_t) + \alpha \mathcal{H}(\pi(\cdot | s_t))\right],
\end{equation}
where $\mathcal{H}$ is the policy entropy and $\alpha$ controls the trade-off between reward and entropy. In this work, we extend SAC with intrinsic rewards derived from expert demonstrations, allowing the agent to benefit from both reward-free guidance and extrinsic rewards when available.

\section{Methods}

This work introduces Mixture of Experts
Guidance using Unlabeled and Incomplete Data for
Exploration (MoE-GUIDE), a novel framework for RL that learns representations from unlabeled data while addressing the challenges posed by the limited information available in expert demonstrations, specifically the presence of gaps in the data and lack of access to actions and next states. These limitations make it infeasible to rely on the conventional techniques, such as demo replay buffers, and leveraging BC. However, expert demonstrations offer valuable insights into desirable trajectories within the environment, even if they are imperfect or limited in scope. We convert these demonstrations into an intrinsic reward, which guides the agent to regions the expert has likely visited. This intrinsic reward can be used alone or as an exploration bonus, allowing the agent to deviate from expert behavior when discovering higher extrinsic rewards. The environment’s reward can prevent the agent from becoming confined to suboptimal behaviors, and a decay function can gradually reduce the influence of expert demonstrations over time. Importantly, since the intrinsic reward is a function of state only, its inclusion does not alter the set of optimal policies for the original environment reward \citep{ng1999policy}. This ensures that while the agent benefits from guided exploration early in training, the optimal solution with respect to the environment’s objective remains unchanged if the intrinsic reward is decayed to 0. We provide a formal argument in Appendix~A.

Pretraining the agent by resetting the simulator to states from expert demonstrations exposes it to regions of the environment visited by the expert, making it easier for the agent to discover and revisit promising areas during training; however, this technique relies on the simulator supporting such resets, which may not be possible in environments with image-based observations or partial observability, and with suboptimal demonstration you may guide the agent to unwanted local minima.

\begin{figure}
    \centering
    \includegraphics[width=0.74\linewidth]{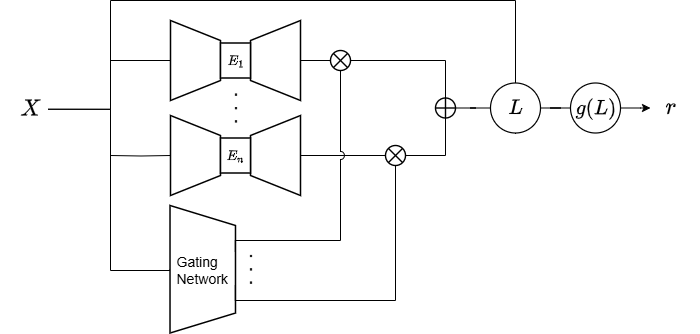}
    \caption{Diagram of the Mixture of Experts framework structure, consisting of \( n \) experts and a gating network. The gating network dynamically assigns weights to experts based on the input state \(X\), enabling collaborative reconstruction of the input through a weighted combination of active experts' outputs. The reconstruction loss \(L\) is normalized and converted into a reward signal \(r\) by the mapping function \(g(L)\), which is then used for guidance.}
    \label{fig:model}
\end{figure}

In this research, we utilize autoencoders as a similarity model, as their bottleneck enables the effective detection of expert behavior, focusing solely on extracting useful features to optimize the reconstruction of expert behavior. We chose autoencoders over variational autoencoders (VAEs) because our tests showed that autoencoders were better at distinguishing expert behavior from other trajectories, while VAEs, likely due to their probabilistic nature, generalized too much and struggled to separate expert from non-expert behavior. To model expert demonstrations, we employ a mixture of autoencoder experts (MoE) as shown in Figure \ref{fig:model}. Rather than relying on the reconstruction loss of a single autoencoder, the experts collectively reconstruct the input as accurately as possible. The MoE model includes two main components: a set of autoencoders (experts) and a gating network, which dynamically assigns a weight to each expert’s output, allowing each autoencoder to specialize in distinct features or patterns of expert behavior.

The final reconstruction is computed as a weighted sum of the outputs of all active experts. For a given input \( x \), each expert \( we\) produces a reconstruction \( \text{expert}_i(x) \), and the gating network assigns a weight \( \text{weight}_i(x) \) to that expert. The final reconstruction \( \hat{x} \) is then given by:
\begin{equation}
    \hat{x} = \sum_{i=1}^{N} \text{weight}_i(x) \cdot \text{expert}_i(x),
\end{equation}

where \( N \) is the number of experts. By enabling the experts to specialize and collaborate, the MoE effectively captures the diverse characteristics of the expert behavior.

To guide exploration, we convert the loss induced by the similarity model into an intrinsic reward signal for the agent. Specifically, we define a mapping function \(g\) that transforms the reconstruction loss at each state, denoted by $L$, into a normalized reward within $[0, 1]$. This process effectively translates the structure of the loss landscape, reflecting the agent's similarity to expert-like states, into intrinsic motivations that can complement the environment’s extrinsic reward.

For a given reconstruction loss $L$, the mapping function $g(L)$ assigns a reward of $1$ when the loss is below a minimum threshold $L_{\min}$, and a reward of $0$ when the loss exceeds a maximum threshold $L_{\max}$. In between the values are normalized between $0$ and $1$, and a monotonically increasing function $f : [0, 1] \rightarrow \mathbb{R}$ is applied, which determines how fast the rewards drop off. The mapping function is defined as
\begin{equation}
    g(L) = \kappa \cdot
 \mathrm{clip}\left( f\left(\dfrac{L - L_{\min}}{L_{\max} - L_{\min}} \right), 0, 1 \right),
\end{equation}
where \(\kappa\) is a scaling factor. In this research, we use an exponential function,
\begin{equation}
    f(x) = e^{-sx},
\end{equation}

where $s$ is a steepness parameter controlling how sharply the reward increases as the loss approaches $L_{\min}$. This mapping provides high rewards for being close to $L_{\min}$ but extremely low rewards when close to $L_{\max}$.

The exploration bonus can, for example, be integrated into the Q-function update equation as follows:
\begin{equation}
    Q'(s, a) = Q(s, a) + \alpha \left( r + \gamma \max_{a'} Q(s', a') - Q(s, a) \right) + \beta \cdot R_{\text{sim}}(s),
\end{equation}
where \(a\) is a given action, \(s\) is a state, \( \alpha \) is the learning rate, \( \gamma \) is the discount factor, and \( R_{\text{sim}}(s) \) is the intrinsic reward derived from the loss. Since \( R_{\text{sim}}(s) \) stays unchanged during the training process, it can be calculated once and added to the replay buffer, limiting the computational overhead of this method. The parameter \( \beta \) controls the influence of the intrinsic reward and can decay over time according to a predefined schedule, such as:
\begin{equation}
    \beta_t = \beta_0 \cdot e^{-\lambda t},
\end{equation}

where \( \beta_0 \) is the initial value of \( \beta \), \( \lambda \) is the decay rate, and \( t \) is the training step. This decay mechanism ensures that the agent relies more heavily on the expert demonstrations during the early stages of training, gradually shifting toward autonomous learning as training progresses. 

\section{Experiments}

We evaluate MoE-GUIDE combined with Soft Actor-Critic (SAC) on five MuJoCo continuous control benchmarks: Swimmer, Hopper, Walker2d, HalfCheetah, and Ant. Each environment is provided with a limited set of expert demonstrations: one for Swimmer, four for Hopper, and ten for Walker2d, HalfCheetah, and Ant. Demonstrations are sparsely sampled by recording states every five steps, resulting in incomplete coverage. The results are averaged over five random seeds, and the shaded areas in the figures represent the standard deviation. Additional details on data collection, environment setup, and hyperparameters, as well as tables listing the final mean rewards for each figure, are provided in the Appendix.

\subsection{Main Evaluation Results}

We compare the following approaches using the average mean reward over 100 episodes: (1) using only extrinsic rewards (ER-only), (2) combining extrinsic rewards with pretraining on demonstration data without having guidance afterwards (ER+pretraining), (3) using the intrinsic reward from the MoE-GUIDE model with pretraining (IR+pretraining), and (4) combining extrinsic and intrinsic rewards (MoE-GUIDE), with pretraining used where applicable. For completeness, we also evaluated RND and ICM baselines; however, these methods performed poorly in dense reward environments, and their results are reported in the Appendix. We evaluated MoE-GUIDE in three settings: perfect expert, imperfect expert, and using sparse rewards.

\begin{figure}[h!] \centering \includegraphics[width=0.99\linewidth]{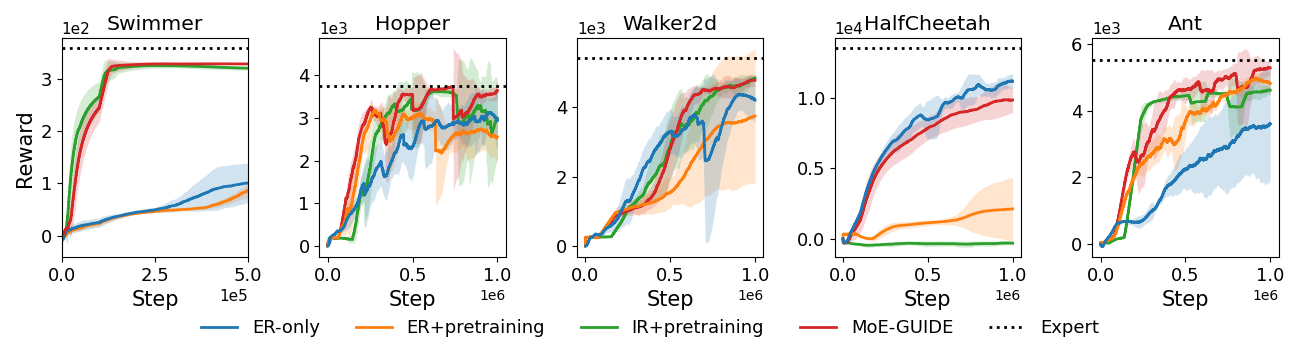} \caption{A comparison of (1) learning with only extrinsic rewards (ER-only), (2) combining extrinsic rewards with pretraining (ER+pretraining), (3) using only the intrinsic reward with pretraining (IR+pretraining), and (4) combining extrinsic and intrinsic rewards (MoE-GUIDE) using expert behavior that achieves high rewards.} \label{fig:perfect} \end{figure}

\textbf{Perfect expert:} Figure~\ref{fig:perfect} compares all methods using strong expert demonstrations. In Swimmer, Walker2d, and Ant, MoE-GUIDE reliably improves over using only extrinsic rewards. In Hopper, both IR-only and MoE-GUIDE reach expert-level performance during training, demonstrating that the intrinsic reward provides particularly effective guidance in this environment. By contrast, for HalfCheetah, extrinsic rewards already effectively guide exploration toward regions yielding high rewards, making it challenging to further improve performance by adding intrinsic rewards.

\begin{figure}[h!] \centering \includegraphics[width=0.89\linewidth]{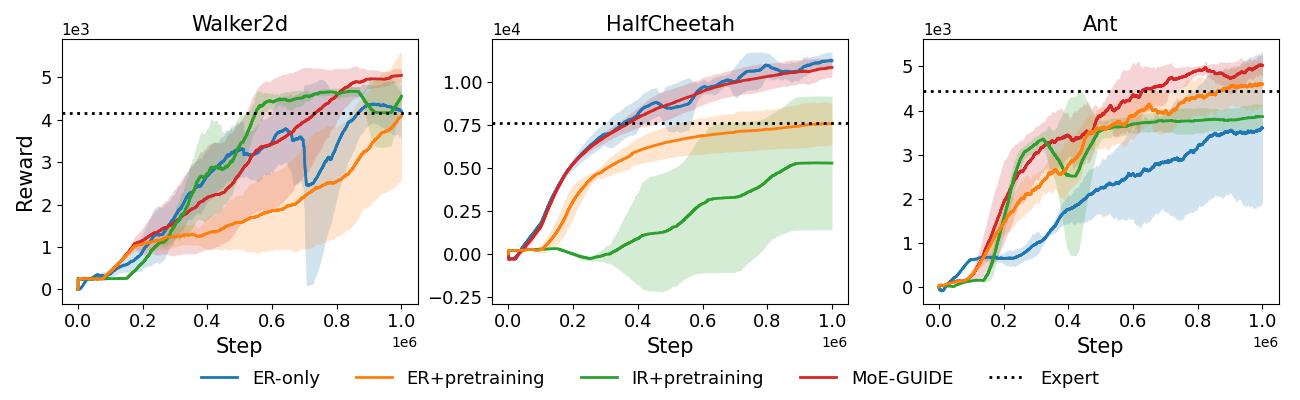} \caption{A comparison of (1) learning with only extrinsic rewards (ER-only), (2) combining extrinsic rewards with pretraining (ER+pretraining), (3) using only the intrinsic reward with pretraining (IR+pretraining), and (4) combining extrinsic and intrinsic rewards (MoE-GUIDE) using imperfect expert behavior.} \label{fig:imp} \end{figure}

\textbf{Imperfect expert:} Figure~\ref{fig:imp} shows results using suboptimal demonstrations. MoE-GUIDE consistently improves over the imperfect expert in all environments. For HalfCheetah, we specifically used a very poor-performing expert to investigate whether leveraging even low-quality demonstrations could still provide value. While MoE-GUIDE did not outperform the extrinsic reward baseline in this case, it achieved comparable and notably more stable results.

\begin{figure}[h!] \centering \includegraphics[width=0.9\linewidth]{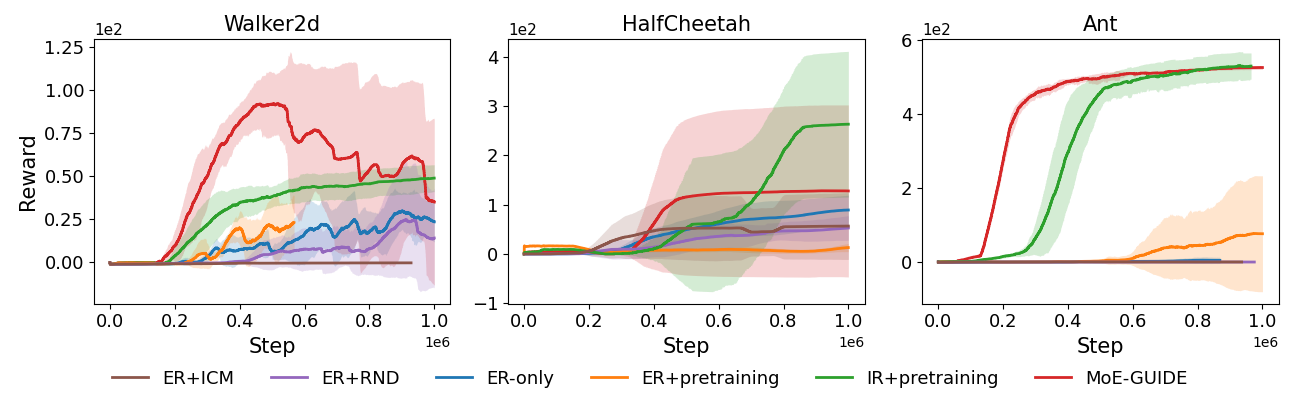} \caption{A comparison of (1) learning with only extrinsic rewards (ER-only), (2) learning from extrinsic rewards combined with intrinsic rewards from ICM or RND, (3) combining extrinsic rewards with pretraining (ER+pretraining), (4) using only the intrinsic reward with pretraining (IR+pretraining), and (5) combining extrinsic and intrinsic rewards (MoE-GUIDE) in a sparse partially observable environment.} \label{fig:sparse} \end{figure}

\textbf{Sparse rewards:} Figure~\ref{fig:sparse} presents results in modified MuJoCo environments where agents only receive rewards for reaching checkpoints at fixed intervals, making exploration more challenging. The agent’s current position (x-coordinate) was omitted from observations and demonstrations for reusability of demonstrations from past experiments, which do not include this, resulting in partial observability and increased task difficulty. This setup also reflects real-world scenarios where state information may be missing due to sensor limitations or the ability to fully observe the environment. Despite never crossing checkpoints during pretraining, MoE-GUIDE significantly outperforms baselines in these sparse, partially observable environments. However, for the HalfCheetah environment, extrinsic rewards benefit MoE-GUIDE early in training, but agents optimizing only intrinsic rewards eventually outperform it.

\textbf{Discussion:} overall, our results demonstrate that MoE-GUIDE consistently achieves strong performance across different environments and expert types, outperforming baseline methods in most cases. The combination of extrinsic and intrinsic rewards is especially beneficial in environments where extrinsic signals are sparse or incomplete.

A more detailed analysis reveals that in dense-reward environments such as HalfCheetah, the extrinsic reward alone effectively guides the agent, leaving limited room for improvement by adding intrinsic motivation. Additionally, our model identifies expert-like behavior near the initial states but struggles to extend guidance beyond this region, causing the agent to remain close to the start. Lowering the $L_{max}$
 threshold can remove some well-recognized initial states, but this further increases the gap to the next expert-like region. To address this, additional intrinsic motivation could be introduced, such as episodic novelty bonuses or other intrinsic signals that encourage exploration beyond early states.

Additionally, we observe that pretraining on demonstration data is a powerful tool; however, without guidance afterwards, it may hinder the learning process of the agent.

In summary, MoE-GUIDE is robust to imperfect demonstrations and excels in environments with sparse or partial reward signals, while its advantage is less pronounced in dense reward environments where extrinsic rewards alone suffice.

\subsection{Ablation Studies}

This section presents an ablation study examining the impact of the number of experts, the gaps in demonstrations, the number of demonstrations, different decay rates, and the mapping function sensitivity for intrinsic reward guidance on the performance of MoE-GUIDE.

\subsubsection{Number of experts}
\begin{figure}[h!]
    \centering
    \includegraphics[width=0.9\linewidth]{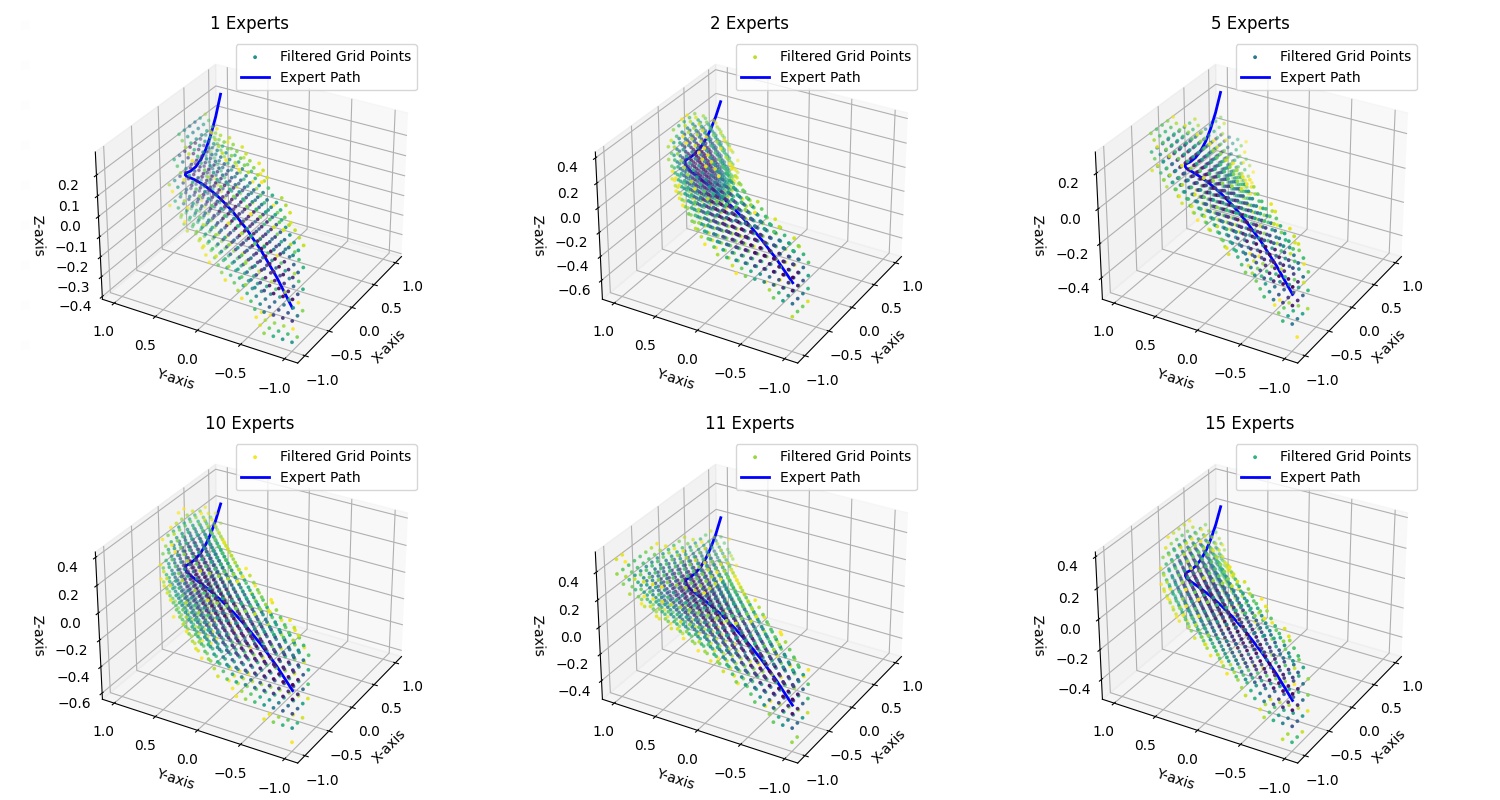}
\caption{Visualization of the loss landscape in the 3D grid world environment. The heatmaps show the loss values at fixed intervals after applying a linear transformation with $loss_\mathrm{max}$. Results are presented for varying numbers of experts, illustrating both the enhanced capacity to identify expert behavior, visible by a purple region around the expert path, and the increased risk of misclassification regions as the number of experts increases.}    \label{fig:experts}
\end{figure}

We introduce a 3D grid world environment to visualize how our model learns from an expert path. The loss landscape is shown as in Figure \ref{fig:experts} with heatmaps indicating the loss at fixed intervals. To highlight model behavior around the expert path, we apply a linear transformation to the loss data using a predefined $L_\mathrm{max}$. In this controlled setting, we perform ablation studies on the number of experts. The results illustrate that increasing the number of experts improves the model's ability to detect expert behavior, shown by a condensed expert-like region with purple around the expert path. However, this also increases the risk of misclassifying other areas as expert-like, as can be seen with 5 and 11 experts, where regions farther from the path are recognized as expert. Notably, the most significant improvement is typically observed when increasing from one to two experts. Based on these findings, we focus on using a low number of experts in our main experiments.

\subsubsection{Demonstration sparsity}

We investigate the impact of demonstration sparsity by varying the number of demonstration episodes and the interval between recorded points, reducing the total number of available samples. As the demonstrations become sparser, we observe a general slight decline in agent performance; however, our method continues to provide meaningful exploration guidance even in these challenging scenarios. Notably, the agents using a single demonstration did not employ pretraining, and our approach still demonstrates robustness in this scenario, with significant gaps resulting in very few demonstration points. It can outperform baselines that rely solely on extrinsic rewards. These results highlight the effectiveness of our method in leveraging even highly limited or imperfect demonstration data to improve exploration.

\begin{figure}[h!]
    \centering
    \includegraphics[width=0.78\linewidth]{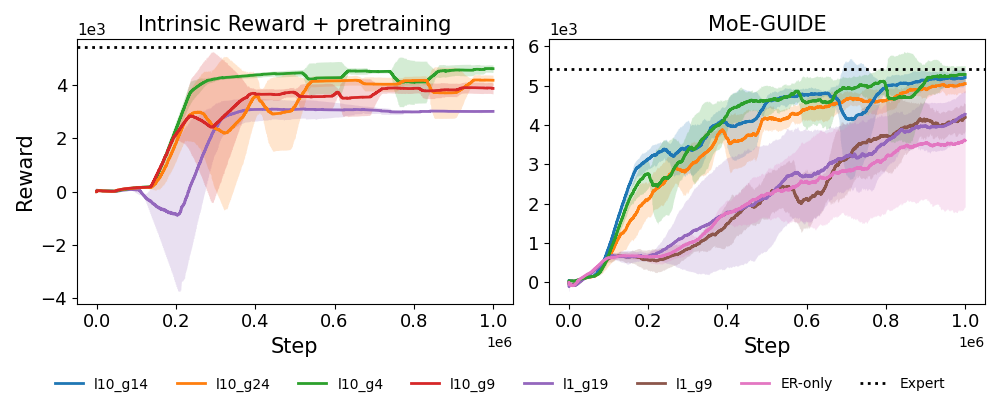}
    \caption{Demonstration size and gap robustness comparison in the Ant environment by varying both the number of demonstration episodes provided to the model, denoted as \( l \), and the gap parameter \( g \), which controls how many samples are skipped between recorded demonstration points. Notably, the agents using a single demonstration did not employ pretraining.}
    \label{fig:slice}
\end{figure}

\subsubsection{Decay Rates}

\begin{figure}[h!]
    \centering
    \includegraphics[width=0.9\linewidth]{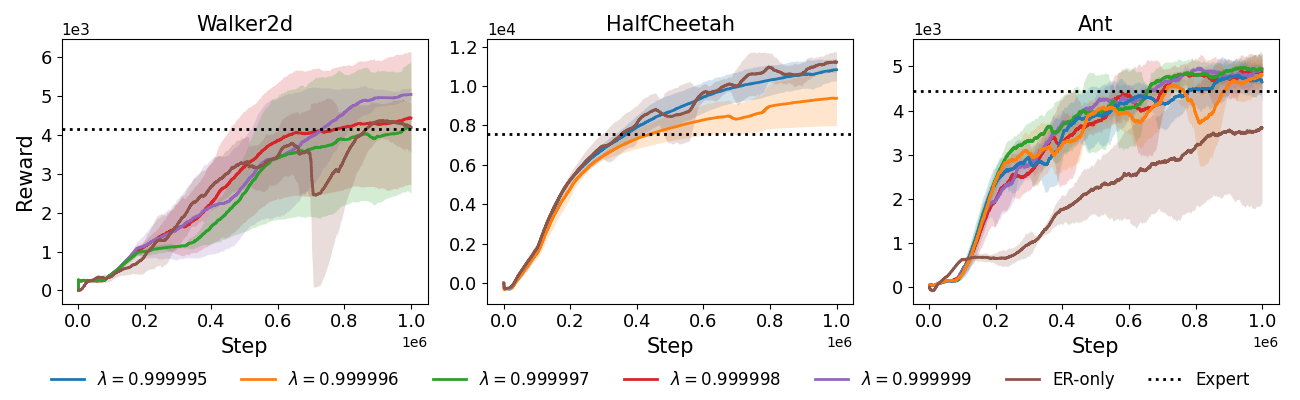}
    \caption{ Learning curves for different intrinsic reward decay rates ($\lambda$). 
        The intrinsic reward is decayed at every timestep using the specified $\lambda$ value.
        A lower decay rate ($\lambda$ closer to 1) maintains expert influence longer, 
        while a higher decay rate quickly reduces the contribution from the expert, 
        allowing the agent to rely more on its own learning.}
    \label{fig:decay}
\end{figure}

In Figure~\ref{fig:decay}, we compare different decay rates for the intrinsic reward. In our implementation, the intrinsic reward is decayed at every time step, so the specified $\lambda$ value is applied at each step, starting from 1. For Walker2d and Ant, a lower decay rate allows the expert’s influence to persist longer during training, resulting in learning behavior and performance that is close to that of a standard agent. In contrast, a higher decay rate is preferred for cases such as HalfCheetah, where the expert performs significantly worse. This enables the agent to benefit from the imperfect expert primarily in the early stages, before quickly transitioning to rely on its own learned policy. Decay rates significantly impact the agent's performance; however, by leveraging intuition about the known strength of the expert behavior, they can be estimated accurately. For instance, in the case of HalfCheetah, we knew we had a very weak expert, so high decay rates were chosen. In contrast, Walker2d performed slightly worse than an average extrinsic reward-only agent, thus requiring a low decay rate.

\subsubsection{Mapping function sensitivity}
\begin{figure}[h!]
    \centering
    \includegraphics[width=0.9\linewidth]{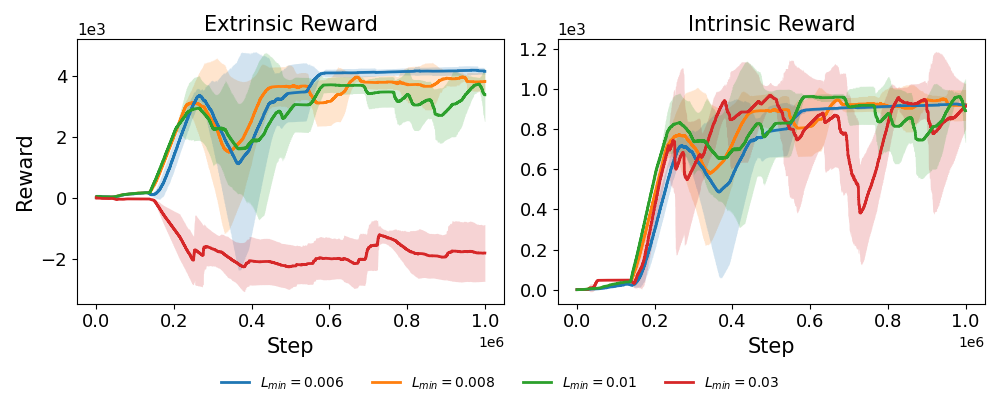}
    \caption{Impact of the mapping threshold $L_{\min}$ on MoE-GUIDE’s performance in the Ant environment. 
High thresholds (e.g., $L_{\min}=0.3$) cause the agent to misidentify many states as expert-like, resulting in high intrinsic but low extrinsic rewards. 
Lower thresholds ($0.01$, $0.008$, $0.006$) yield better extrinsic returns, though setting $L_{\min}$ too low can make it harder for the agent to find expert-like regions after pretraining.}
    \label{fig:loss}
\end{figure}

The choice of mapping function thresholds is critical for MoE-GUIDE's performance, as shown by different $L_{\min}$ in Figure~\ref{fig:loss}. High $L_{\min}$ (e.g., 0.3) leads the agent to treat many states as expert-like, resulting in suboptimal behavior and low extrinsic rewards, while nearly maximizing intrinsic reward. Lower values (0.01, 0.008, 0.006) yield strong extrinsic performance, with the lowest best overall, though setting $L_{\min}$ too low can prevent the agent from finding expert-like regions after pretraining. The optimal threshold depends on whether pretraining is used, since expert-like regions can be narrower when the agent visits them during pretraining. Although the tresholds are sensitive, it is straightforward to diagnose poor choices: if the model represents random trajectories well, $L_{\min}$ or $L_{\max}$ is likely too large and if expert behavior is not captured, the values are too low.

\section{Conclusion \& future works}
We present MoE-GUIDE, a method for directed exploration in Reinforcement Learning that uses a mixture of similarity models trained on expert demonstrations to construct a loss landscape, which resembles the similarity of each state to expert behavior. This loss is then transformed into an intrinsic reward through a mapping function, guiding the agent towards expert-like states. MoE-GUIDE is effective in both dense and sparse reward environments, demonstrating versatility across a range of exploration challenges. The method operates successfully with demonstrations that are unlabeled, incomplete, or imperfect. Our results show that agents benefit from MoE-GUIDE, even with limited data that contains gaps. One limitation of our method is the need for manual selection of the similarity model and mapping function. In environments where extrinsic rewards already provide sufficient guidance, the additional intrinsic reward may be less beneficial and can lead to suboptimal exploration. Future work could incorporate episodic intrinsic motivation to prevent the agent from repeatedly visiting similar states and explore alternative or adaptive similarity models. In this work, we have only tested autoencoders and variational autoencoders; however, other methods, such as density estimation, ICM, and RND, could also be considered for future research. Additionally, being able to inspect the loss landscape for expert-behavior representation and misclassifications would enable more efficient and effective mapping functions.

\appendix

\section{Properties and pitfalls of state-only intrinsic motivation}

\subsection{Proof that state-only intrinsic motivation does not change the optimal policy}

Let $r_{\text{env}}(s,a)$ denote the environment (extrinsic) reward, and let $r_{\text{int}}(s)$ denote an intrinsic reward that depends only on the state $s$. The agent receives the total reward:
\[
r_{\text{total}}(s, a) = r_{\text{env}}(s, a) + r_{\text{int}}(s).
\]

Let $V^{\pi}_{\text{env}}(s)$ be the value function under policy $\pi$ and reward $r_{\text{env}}$, and $V^{\pi}_{\text{total}}(s)$ under $r_{\text{total}}$:
\[
V^{\pi}_{\text{env}}(s) = \mathbb{E}_{\pi}\left[ \sum_{t=0}^\infty \gamma^t r_{\text{env}}(s_t, a_t) \mid s_0 = s \right],
\]
\[
V^{\pi}_{\text{total}}(s) = \mathbb{E}_\pi\left[ \sum_{t=0}^\infty \gamma^t (r_{\text{env}}(s_t, a_t) + r_{\text{int}}(s_t)) \mid s_0 = s \right].
\]

Expanding $V^{\pi}_{\text{total}}(s)$, we get:
\begin{align*}
V^{\pi}_{\text{total}}(s) &= \mathbb{E}_\pi\left[ \sum_{t=0}^\infty \gamma^t r_{\text{env}}(s_t, a_t) + \sum_{t=0}^\infty \gamma^t r_{\text{int}}(s_t) \mid s_0 = s \right] \\
&= V^{\pi}_{\text{env}}(s) + V^{\pi}_{\text{int}}(s),
\end{align*}
where
\[
V^{\pi}_{\text{int}}(s) = \mathbb{E}_\pi\left[ \sum_{t=0}^\infty \gamma^t r_{\text{int}}(s_t) \mid s_0 = s \right].
\]

The set of optimal policies under $r_{\text{env}}$ is
\[
\pi^*_{\text{env}} = \arg \max_\pi V^{\pi}_{\text{env}}(s), \quad \forall s.
\]
Under $r_{\text{total}}$,
\[
\pi^*_{\text{total}} = \arg\max_\pi V^{\pi}_{\text{total}}(s).
\]

\textbf{Observation:} The difference in value between any two policies $\pi_1$ and $\pi_2$ is the same under $V^{\pi}_{\text{env}}$ and $V^{\pi}_{\text{total}}$:
\[
V^{\pi_1}_{\text{total}}(s) - V^{\pi_2}_{\text{total}}(s) = \left(V^{\pi_1}_{\text{env}}(s) - V^{\pi_2}_{\text{env}}(s)\right) + \left(V^{\pi_1}_{\text{int}}(s) - V^{\pi_2}_{\text{int}}(s)\right)
\]

However, $V^{\pi}_{\text{int}}(s)$ depends only on the state visitation distribution induced by $\pi$. Since $r_{\text{int}}(s)$ does not depend on actions, optimizing $V^{\pi}_{\text{total}}(s)$ is equivalent to optimizing $V^{\pi}_{\text{env}}(s)$, as $V^{\pi}_{\text{int}}(s)$ is additive and does not affect the relative ordering of policies with respect to $V^{\pi}_{\text{env}}(s)$.

\textbf{Conclusion:} The set of optimal policies is unchanged. That is,
\[
\Pi^*_{\text{env}} = \Pi^*_{\text{total}}.
\]
Thus, adding a state-only intrinsic reward does not alter the optimal policy for the original environment reward.

This result follows the classic reward shaping theory as discussed in \citet{ng1999policy}. For completeness, we reproduce the argument here.

\begin{flushright}
$\Box$
\end{flushright}

\subsection{Illustrative example of possible pitfalls}
We now show an example of a pitfall of using state-only intrinsic motivation. When intrinsic rewards depend solely on visiting specific states, the agent can become overly focused on those states that provide intrinsic reward, rather than exploring the environment for potentially higher extrinsic rewards. This can lead to undesirable behavior where the agent repeatedly visits or remains within rewarding states, effectively becoming ``stuck'' in these regions. As a result, the agent may fail to discover more optimal strategies or reach states with significant extrinsic rewards.

If the observation includes velocity or frame-stacked states, the agent is still incentivised to move, as trying to match the velocity of the expert will encourage the agent to traverse the environment rather than remain in a single region.

Table~\ref{tab:rewards} details the intrinsic and extrinsic rewards for each state in an example Markov Decision Process (MDP). The agent receives an intrinsic reward of $+1$ for visiting states S$_1$, S$_2$, and S$_6$. The terminal state S$_6$ also provides a larger extrinsic reward of $+10$. Table~\ref{tab:actions} explains the possible actions.

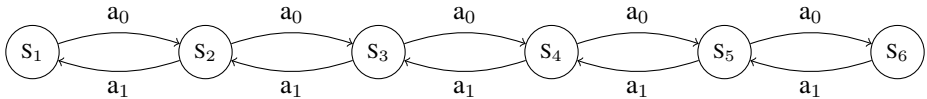
\begin{figure}[H]
\centering
\begin{tikzpicture}[
  node distance=2.3cm, 
  every state/.style={minimum size=18pt, font=\footnotesize},
  curlyarrow/.style={->, decorate, decoration={snake, segment length=7pt, amplitude=1.2pt}, thick}
  ]

  \node[state] (S1) {S$_1$};
  \node[state, right of=S1] (S2) {S$_2$};
  \node[state, right of=S2] (S3) {S$_3$};
  \node[state, right of=S3] (S4) {S$_4$};
  \node[state, right of=S4] (S5) {S$_5$};
  \node[state, right of=S5] (S6) {S$_6$};

  \draw[->] (S1) to[bend left=18] node[above] {a$_0$} (S2);
  \draw[->] (S2) to[bend left=18] node[above] {a$_0$} (S3);
  \draw[->] (S3) to[bend left=18] node[above] {a$_0$} (S4);
  \draw[->] (S4) to[bend left=18] node[above] {a$_0$} (S5);
  \draw[->] (S5) to[bend left=18] node[above] {a$_0$} (S6);

  \draw[->] (S2) to[bend left=18] node[below] {a$_1$} (S1);
  \draw[->] (S3) to[bend left=18] node[below] {a$_1$} (S2);
  \draw[->] (S4) to[bend left=18] node[below] {a$_1$} (S3);
  \draw[->] (S5) to[bend left=18] node[below] {a$_1$} (S4);
  \draw[->] (S6) to[bend left=18] node[below] {a$_1$} (S5);

\end{tikzpicture}
\caption{MDP transition graph. States S$_1$ to S$_6$ are connected by curly arrows denoting actions a$_0$ (right) and a$_1$ (left).}
\label{fig:mdp-graph}
\end{figure}

Figure~\ref{fig:mdp-graph} shows the transition structure of this environment. At each state, the agent can execute action $a_0$ (move right) or $a_1$ (move left). If the immediate intrinsic reward primarily drives the agent's policy, it may become trapped, oscillating between the early rewarding states (S$_1$, S$_2$, S$_3$), and fail to reach the high extrinsic reward at S$_6$.

\begin{table}[H]
\caption{Intrinsic and extrinsic rewards for each state in the environment.}
\label{tab:rewards}
\centering
\begin{tabular}{ccc}
\toprule
\textbf{State} & \textbf{Intrinsic Reward} & \textbf{Extrinsic Reward} \\
\midrule
S$_1$ & $+1$ & $0$ \\
S$_2$ & $+1$ & $0$ \\
S$_3$ & $+1$  & $0$ \\
S$_4$ & $0$  & $0$ \\
S$_5$ & $0$ & $0$ \\
S$_6$ & $+1$ & $+10$ \\
\bottomrule
\end{tabular}
\end{table}

\begin{table}[H]
\caption{Action meanings in the MDP.}
\label{tab:actions}
\centering
\begin{tabular}{ll}
\toprule
\textbf{Action} & \textbf{Description} \\
\midrule
a$_0$ & Move one state to the right \\
a$_1$ & Move one state to the left \\
\bottomrule
\end{tabular}
\end{table}

To mitigate this, intrinsic reward schemes can be enhanced in several ways:
\begin{itemize}
    \item \textbf{Global intrinsic rewards decay}: After every episode or time step, \(\beta\) can be decayed so that the agent focuses more on maximizing the extrinsic rewards over time.
    \item \textbf{State/region-specific intrinsic rewards decay:} A more advanced way of decaying the intrinsic reward is to decay the intrinsic rewards for a specific state or region either within an episode (episodic novelty) or across all episodes (lifetime novelty), reducing the incentive to revisit familiar states, while not decaying the intrinsic rewards for unvisited states.
    \item \textbf{Exploration Bonuses:} Methods such as Random Network Distillation (RND) or Intrinsic Curiosity Module (ICM) provide additional motivation for exploration by rewarding the agent for encountering novel or unpredictable states.
\end{itemize}

Careful design of intrinsic motivation is crucial to avoid behaviors where agents are incentivized to remain in suboptimal regions, thereby detracting from overall task performance.

\section{Experimental settings}

\subsection{Soft Actor-Critic}

\begin{table}[h!]
\centering
\caption{Key hyperparameters for Soft Actor-Critic (SAC) in RLlib. All settings are default unless noted. Replay buffer size for Swimmer is 100{,}000; for all other environments, it is 1{,}000{,}000.}
\begin{tabular}{ll}
\toprule
\textbf{Parameter} & \textbf{Value} \\
\midrule
Discount factor ($\gamma$) & 0.99 \\
Actor learning rate & 0.0003 \\
Critic learning rate & 0.0003 \\
Entropy learning rate & 0.0003 \\
Optimizer & Adam \\
Target smoothing coefficient ($\tau$) & 0.005 \\
Target network update frequency & 0 \\
Replay buffer size & 1,000,000 (100,000 for Swimmer) \\
Batch size & 256 \\
Number of hidden layers & 2 \\
Hidden layer size & 256 \\
Activation function & ReLU \\
Target entropy & auto \\
N-step returns & 1 \\
Action normalization & True \\
\bottomrule
\end{tabular}
\label{tab:sac_hyperparameters}
\end{table}

\subsection{Mixture of autoencoders hyperparameters and training details}

Tables~\ref{tab:main-hyperparams}--\ref{tab:gaps-table} present all the hyperparameters for the models used in Section~4 of this paper. These tables comprehensively document the configuration for each environment and model variant (Table~\ref{tab:main-hyperparams}), the parameters for the sparse environment  (Table~\ref{tab:sparse-hyperparams}), the decay parameter settings (Table~\ref{tab:decay-table}), additional configurations for iterative mask pruning (IMP) experiments (Table~\ref{tab:imp-table}), the loss values explored (Table~\ref{tab:loss-table}), and ablations on gap hyperparameters (Table~\ref{tab:gaps-table}). 

All models are trained using the Adam optimizer with a learning rate of 0.001 for 3000 epochs. The choice of 3000 training epochs is motivated by the need to balance the models' ability to closely fit the expert demonstrations with their ability to generalize to unseen states. We observed that increasing the number of training epochs consistently decreased the reconstruction error on the training data; however, excessively long training can reduce the model's ability to generalize, as it may overfit to the expert data. Thus, 3000 epochs were selected as a compromise between accurate representation of the expert trajectories and generalization performance.

\begin{table}[h!]
\centering
\caption{Hyperparameter configurations for the perfect agents experiment for each environment and model variant.}
\label{tab:main-hyperparams}
\scriptsize
\setlength{\tabcolsep}{3pt}
\renewcommand{\arraystretch}{0.95}
\begin{tabularx}{0.98\textwidth}{l l Y Y Y Y Y Y Y}
\toprule
\textbf{Environment} & \textbf{Model} & \textbf{Bottleneck} & \textbf{Num Experts} & $\mathbf{L_{min}}$ & $\mathbf{L_{max}}$ & \textbf{Mapping Function} & \textbf{Steepness} & \textbf{Scale factor} \\
\midrule
Swimmer      & MoE-GUIDE        & 3  & 1 & 0.01    & 0.1  & Exponential & 20  & 1  \\
             & IR+pretraining   & 3  & 1 & 0.01    & 0.1  & Exponential & 20  & 1  \\
Hopper       & MoE-GUIDE        & 4  & 2  & 0.03    & 0.05 & Exponential & 100 & 2  \\
             & IR+pretraining   & 4  & 2  & 0.03    & 0.05 & Exponential & 200 & 2  \\
HalfCheetah  & MoE-GUIDE        & 7  & 4 & 0.1     & 0.9  & Exponential & 100 & 1 \\
             & IR+pretraining   & 7  & 4 & 0.1     & 0.9  & Exponential & 200 & 1 \\
Walker2d     & MoE-GUIDE        & 7  & 3 & 0.04    & 0.5  & Exponential & 100 & 2 \\
             & IR+pretraining   & 7  & 3 & 0.04    & 0.5  & Exponential & 200 & 2 \\
Ant          & MoE-GUIDE        & 10 & 2  & $4\times10^{-5}$ & 0.1  & Exponential & 100 & 2 \\
             & IR+pretraining   & 10 &  2 & $4\times10^{-5}$ & 0.1  & Exponential & 200 & 2 \\
\bottomrule
\end{tabularx}
\end{table}

\begin{table}[h!]
\centering
\caption{Hyperparameter configurations for the agents trained in the sparse environment for each environment and model variant.}
\label{tab:sparse-hyperparams}
\scriptsize
\setlength{\tabcolsep}{3pt}
\renewcommand{\arraystretch}{0.95}
\begin{tabularx}{0.98\textwidth}{l l Y Y Y Y Y Y Y}
\toprule
\textbf{Environment} & \textbf{Model} & \textbf{Bottleneck} & \textbf{Num Experts} & $\mathbf{L_{min}}$ & $\mathbf{L_{max}}$ & \textbf{Mapping Function} & \textbf{Steepness} & \textbf{Scale factor} \\
\midrule

HalfCheetah  & MoE-GUIDE        & 7  & 4 & 0.1     & 0.9  & Exponential & 200 & 0.01 \\
             & IR+pretraining   & 7  & 4 & 0.1     & 0.9  & Exponential & 200 & 1 \\
Walker2d     & MoE-GUIDE        & 7  & 3 & 0.04    & 0.5  & Exponential & 100 & 0.01 \\
             & IR+pretraining   & 7  & 3 & 0.04    & 0.5  & Exponential & 200 & 2 \\
Ant          & MoE-GUIDE        & 10 & 2  & $4\times10^{-5}$ & 0.1  & Exponential & 200 & 0.01 \\
             & IR+pretraining   & 10 &  2 & $4\times10^{-5}$ & 0.1  & Exponential & 200 & 0.01 \\
\bottomrule
\end{tabularx}
\end{table}

\begin{table}[h!]
\centering
\caption{Decay parameter configurations for the decay ablation study for Ant, HalfCheetah, and Walker2d.}
\label{tab:decay-table}
\scriptsize
\setlength{\tabcolsep}{3pt}
\renewcommand{\arraystretch}{0.95}
\begin{tabularx}{0.98\textwidth}{l Y Y Y Y Y Y Y Y Y}
\toprule
Env & Decay & Model & Bottleneck & Experts & $L_{min}$ & $L_{max}$ & Map Fn & Steepness & Scale \\
\midrule
Ant         & 0.999995 & MoE-GUIDE    & 10 & 3 & $6\times10^{-4}$ & 0.01 & Exp & 100 & 5 \\
            & 0.999996 & MoE-GUIDE    & 10 & 3 & $6\times10^{-4}$ & 0.01 & Exp & 100 & 5 \\
            & 0.999997 & MoE-GUIDE    & 10 & 3 & $6\times10^{-4}$ & 0.01 & Exp & 100 & 5 \\
            & 0.999998 & MoE-GUIDE    & 10 & 3 & $6\times10^{-4}$ & 0.01 & Exp & 100 & 5 \\
            & 0.999999 & MoE-GUIDE    & 10 & 3 & $6\times10^{-4}$ & 0.01 & Exp & 100 & 5 \\
HalfCheetah & 0.999995 & MoE-GUIDE    & 7  & 4 & 0.1             & 0.8  & Exp & 100 & 5 \\
            & 0.999996 & MoE-GUIDE    & 7  & 4 & 0.1             & 0.8  & Exp & 100 & 5 \\
Walker2d    & 0.999997 & MoE-GUIDE    & 7  & 4 & 0.03            & 0.5  & Exp & 100 & 5 \\
            & 0.999998 & MoE-GUIDE    & 7  & 4 & 0.03            & 0.5  & Exp & 100 & 5 \\
            & 0.999999 & MoE-GUIDE    & 7  & 4 & 0.03            & 0.5  & Exp & 100 & 5 \\
\bottomrule
\end{tabularx}
\end{table}

\begin{table}[h!]
\centering
\caption{Hyperparameter configurations for the experiment using imperfect experts for Ant, HalfCheetah, and Walker2d.}
\label{tab:imp-table}
\scriptsize
\setlength{\tabcolsep}{3pt}
\renewcommand{\arraystretch}{0.95}
\begin{tabularx}{0.98\textwidth}{l l Y Y Y Y Y Y Y}
\toprule
\textbf{Environment} & \textbf{Model} & \textbf{Bottleneck} & \textbf{Num Experts} & $\mathbf{L_{min}}$ & $\mathbf{L_{max}}$ & \textbf{Mapping Function} & \textbf{Steepness} & \textbf{Scale factor} \\
\midrule
HalfCheetah  & MoE-GUIDE        & 7  & 4 & 0.1   & 0.8  & Exponential & 100 & 1 \\
             & IR+pretraining   & 7  & 4 & 0.1   & 0.8  & Exponential & 200 & 1 \\
Walker2d     & MoE-GUIDE        & 7  & 4 & 0.03  & 0.5  & Exponential & 100 & 2 \\
             & IR+pretraining   & 7  & 4 & 0.03  & 0.5  & Exponential & 200 & 2 \\
Ant & MoE-GUIDE      & 10 & 3 & $6\times10^{-4}$ & 0.01 & Exponential & 100 & 1 \\
    & IR+pretraining & 10 & 3 & $6\times10^{-4}$ & 0.01 & Exponential & 200 & 1 \\
\bottomrule
\end{tabularx}
\end{table}

\begin{table}[h!]
\centering
\caption{Max loss values explored in the Ant environment, with all hyperparameter columns.}
\label{tab:loss-table}
\scriptsize
\setlength{\tabcolsep}{3pt}
\renewcommand{\arraystretch}{0.95}
\begin{tabularx}{0.95\textwidth}{l Y Y Y Y Y Y Y}
\toprule
\textbf{Loss Value} & \textbf{Model} & \textbf{Bottleneck} & \textbf{Num Experts} & $\mathbf{L_{max}}$ & \textbf{Mapping Function} & \textbf{Steepness} & \textbf{Scale factor} \\
\midrule
0.01  & IR+pretraining & 6 & 2 & 0.1 & Exponential & 100 & 1 \\
0.03  & IR+pretraining & 6 & 2 & 0.1 & Exponential & 100 & 1 \\
0.006 & IR+pretraining & 6 & 2 & 0.1 & Exponential & 100 & 1 \\
0.008 & IR+pretraining & 6 & 2 & 0.1 & Exponential & 100 & 1 \\
\bottomrule
\end{tabularx}
\end{table}

\begin{table}[h!]
\centering
\caption{Gaps results configurations in the Ant environment, with all hyperparameter columns.}
\label{tab:gaps-table}
\scriptsize
\setlength{\tabcolsep}{3pt}
\renewcommand{\arraystretch}{0.95}
\begin{tabularx}{0.95\textwidth}{l l Y Y Y Y Y Y Y}
\toprule
\textbf{Name} & \textbf{Model} & \textbf{Bottleneck} & \textbf{Num Experts} & $\mathbf{L_{min}}$ & $\mathbf{L_{max}}$ & \textbf{Mapping Function} & \textbf{Steepness} & \textbf{Scale factor} \\
\midrule
l10\_s5   & MoE-GUIDE      & 10 & 2  & $4\times10^{-5}$ & 0.1  & Exponential & 100 & 2 \\
l10\_s5   & IR+pretraining & 10 & 2  & $4\times10^{-5}$ & 0.1  & Exponential & 200 & 2 \\
l10\_s10  & IR+pretraining & 10 & 3  & $8\times10^{-4}$ & 0.1  & Exponential & 200 & 1 \\
l10\_s15  & MoE-GUIDE      & 10 & 3  & $3\times10^{-4}$ & 0.05 & Exponential & 100 & 2 \\
l10\_s25  & MoE-GUIDE      & 10 & 3  & $1\times10^{-4}$ & 0.05 & Exponential & 100 & 2 \\
l1\_s10   & MoE-GUIDE      & 10 & 10 & 0.0001           & 0.08 & Exponential & 100 & 1 \\
l1\_s20   & IR+pretraining & 10 & 5  & 0.01             & 0.05 & Exponential & 200 & 1 \\
\bottomrule
\end{tabularx}
\end{table}

\subsection{RND \& ICM}

For our experiments involving intrinsic motivation, we implemented Intrinsic Curiosity Module (ICM) and Random Network Distillation (RND) as auxiliary reward signals. The design and hyperparameter selection for these methods was guided by the original works \cite{ICM, RND}, as well as by \cite{yuan2024rlexplore}, which provides extensive discussion of architectural choices, normalization strategies, and practical recommendations. In particular, \cite{yuan2024rlexplore} informed our choices of network size, orthogonal initialization, and state normalization. Additionally, following the findings of \cite{li2019curiosity}, we use only the forward model in off-policy settings, as it was shown to be sufficient for effective curiosity-driven exploration.

The following choices and hyperparameters were used consistently for both ICM and RND:

\begin{itemize}
    \item \textbf{State Normalisation:} All states were normalised before being fed to the intrinsic modules.
    \item \textbf{Reward Normalisation:} Intrinsic rewards were normalised online using a running mean and standard deviation (RMS).
    \item \textbf{Optimizer:} Adam optimizer with a learning rate of 0.0003.
    \item \textbf{Feature Dimension:} 128-dimensional feature space for the learned or random embeddings.
    \item \textbf{Hidden Dimension:} All multi-layer perceptrons (MLPs) used in the intrinsic modules had hidden layers of size 256.
    \item \textbf{Weight Initialization:} All networks were initialized using orthogonal initialization.
    \item \textbf{$\beta$:} was chosen to provide a small exploration bonus of around \(~ 1\%\) of the extrinsic reward.
\end{itemize}

\section{Data Source Selection Rationale}

The selection of data sources for both the perfect and imperfect experts was guided by the need to provide a comprehensive evaluation of the agent’s learning capabilities under varying supervision qualities. For the \textbf{perfect experts}, we prioritized the best available baselines, ensuring that the guidance provided to the agents represented near-optimal behavior. This allowed for a stringent assessment of whether agents could, with the aid of such expert data, achieve or approximate top-tier performance within a limited training budget of 1 million timesteps.

For the \textbf{imperfect experts}, we aimed to supply the agents with guidance that, while informative, did not represent optimal behavior. Specifically, we selected agents that performed slightly below the average Soft Actor-Critic (SAC) agent but were not trapped in poor local optima, thereby providing learning signals that were sub-optimal yet still constructive. An exception was made for the HalfCheetah environment, where we intentionally included a particularly poorly performing agent. This choice was made to test whether even data from significantly subpar agents can be rigorously
could contribute positively to the learning process when combined with the proposed guidance.

Together, these selections facilitate a thorough investigation into the robustness and effectiveness of the learning algorithms when exposed to both high-quality and imperfect supervision.

\begin{table}[h!]
\centering
\caption{Data sources used for training models, categorized by expert quality.}
\begin{tabular}{l l l p{5cm}}
\hline
\textbf{Category} & \textbf{Environment} & \textbf{Mean reward} & \textbf{Source/Description} \\
\hline
Perfect Expert & Swimmer & 359.51 & Standard settings Open-Loop Baseline \cite{raffin2023open}  \\
 & Hopper & 3760.69 & CILO paper dataset \cite{gavenski2024explorative} \\
 & Walker2d & 5413.74 & Good performing SAC agent after 2M timesteps  \\
 & HalfCheetah & 13550.97 & Good performing SAC agent after 2M timesteps  \\
 & Ant & 5514.02 & CILO paper dataset \cite{gavenski2024explorative} \\
\hline
Imperfect Expert & Walker2d & 4149.77 & Below average SAC agent after 1M timesteps  \\
 & HalfCheetah & 7581.55 & CILO paper dataset  \cite{gavenski2024explorative} \\
 & Ant & 4442.53 & Slightly below average performing SAC agent  after 1M timesteps that is not stuck in a bad local minima \\
\hline
\end{tabular}

\end{table}

\section{Environment details}

\begin{figure}[h!]
    \centering
    \begin{subfigure}[b]{0.19\textwidth}
        \centering
        \includegraphics[width=\linewidth]{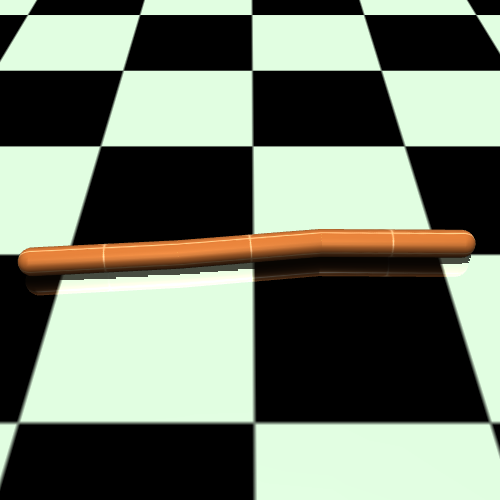}
        \caption{Swimmer}
    \end{subfigure}
    \hfill
    \begin{subfigure}[b]{0.19\textwidth}
        \centering
        \includegraphics[width=\linewidth]{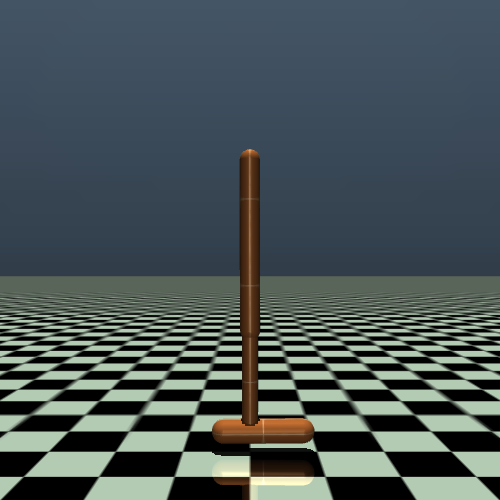}
        \caption{Hopper}
    \end{subfigure}
    \hfill
    \begin{subfigure}[b]{0.19\textwidth}
        \centering
        \includegraphics[width=\linewidth]{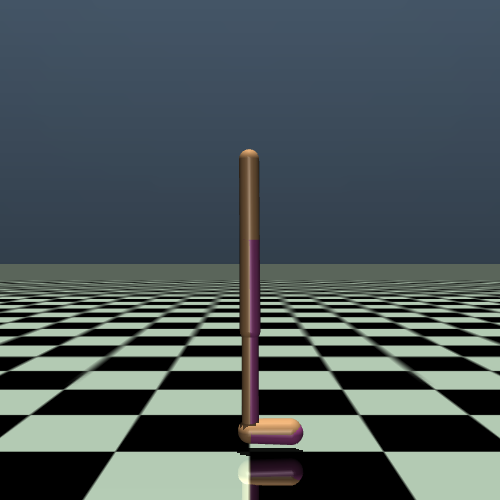}
        \caption{Walker2d}
    \end{subfigure}
    \hfill
    \begin{subfigure}[b]{0.19\textwidth}
        \centering
        \includegraphics[width=\linewidth]{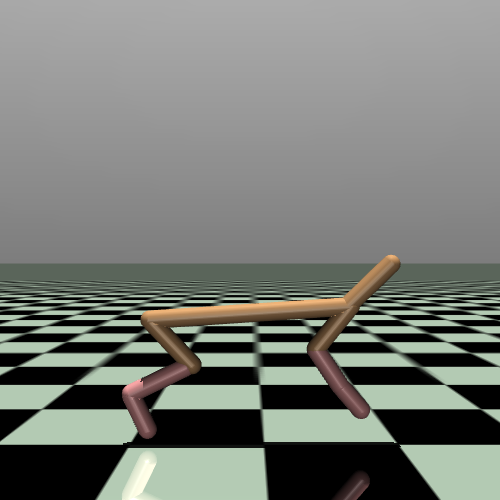}
        \caption{HalfCheetah}
    \end{subfigure}
    \hfill
    \begin{subfigure}[b]{0.19\textwidth}
        \centering
        \includegraphics[width=\linewidth]{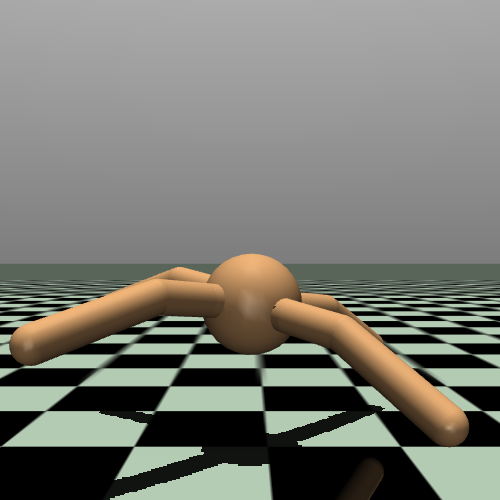}
        \caption{Ant}
    \end{subfigure}
    \caption{Screenshots of the MuJoCo environments used as baselines for locomotion experiments.}
    \label{fig:mujoco_envs}
\end{figure}

We employ five standard continuous control environments from the MuJoCo suite: Swimmer, Hopper, Walker2d, HalfCheetah, and Ant. These environments are widely used to benchmark reinforcement learning algorithms in simulated robotic locomotion. In Table \ref{tab:mujoco_spaces}, details about the action and state spaces can be found, and Figure \ref{fig:mujoco_envs} visualizes the environments.

\begin{table}[h!]
\centering
\caption{Observation and action space dimensions for MuJoCo environments. Here, $\mathcal{S}$ denotes the state (observation) space and $\mathcal{A}$ denotes the action space.}
\begin{tabular}{lcc}
\toprule
\textbf{Environment} & $\dim(\mathcal{S})$ & $\dim(\mathcal{A})$ \\
\midrule
Swimmer-v2      & 8  & 2 \\
Hopper-v2       & 11 & 3 \\
Walker2d-v2     & 17 & 6 \\
HalfCheetah-v2  & 17 & 6 \\
Ant-v2          & 111 & 8 \\
\bottomrule
\end{tabular}
\label{tab:mujoco_spaces}
\end{table}

For our experiments requiring sparse rewards, we modified the MuJoCo environments so that agents receive rewards only upon reaching checkpoints at fixed intervals, 2.5 for Walker2d, 5 for Ant, and 11 for HalfCheetah. These intervals were chosen to ensure that checkpoints are reachable, while still presenting a challenging exploration problem for the agent. Early episode termination results in a total reward of $-1$, incentivizing safe and deliberate exploration. HalfCheetah does not terminate early so this makes the exploration easier.

In the sparse-reward setup, the agent’s position is not included in the state or demonstrations, making the environment partially observable. Additionally, the environment cannot be reset to the agent’s exact demonstration location; instead, the agent is respawned at the initial starting point but initialized with the expert’s joint angles and velocities. This setup exposes the agent to expert behavior without providing immediate extrinsic rewards.

\section{Traditional intrinsic reward methods baseline}
\begin{figure}[h!]
    \centering
    \includegraphics[width=1\linewidth]{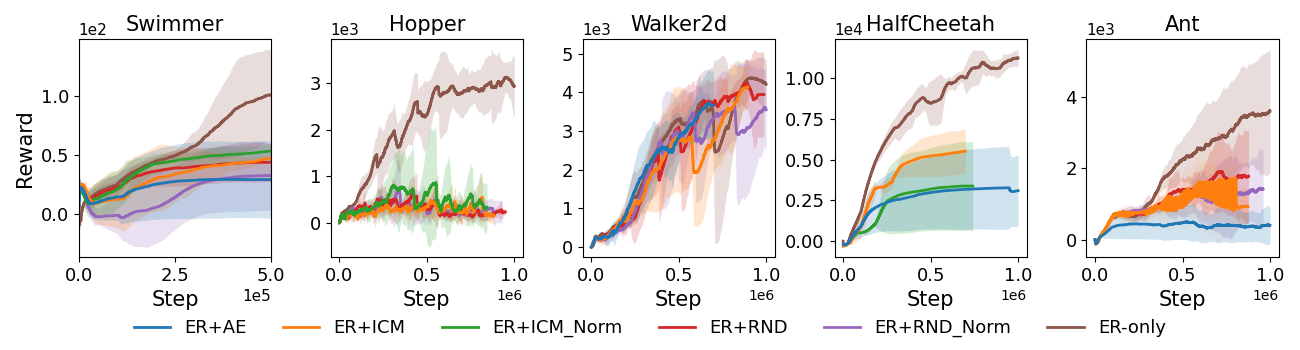}
    \caption{Performance comparison of agents trained with standard intrinsic rewards (ICM, RND, and autoencoder-based) and normalized rewards versus extrinsic rewards in a dense environment. }
    \label{intri}
\end{figure}

In this section, we investigate the impact of various intrinsic motivation methods, using both standard and normalized rewards. Specifically, Intrinsic Curiosity Module (ICM), Random Network Distillation (RND), and an autoencoder-based intrinsic reward, on agent performance in a dense reward environment. While these intrinsic rewards are designed to encourage exploration in sparse settings, we observe in Figure \ref{intri} that in environments with dense rewards, they can hinder the learning process. In Halcheetah, ER+ICM has a strange shape, but this is due to the highly unstable learning curves of all the runs.

\bibliography{main}
\bibliographystyle{rlj}

\beginSupplementaryMaterials
\section{Grid world}
We present qualitative results in a gridworld with random walls, where the agent can move in any direction. The agent always selects randomly among actions that yield the highest intrinsic reward. For our method (MoE-GUIDE), intrinsic rewards are only given once per state to prevent the agent from getting stuck revisiting the same locations. In all visualizations, the green star indicates the start state, the red star marks the goal, and blue dots represent demonstration states.

\begin{figure} [h!]
    \centering
    \includegraphics[width=1\textwidth]{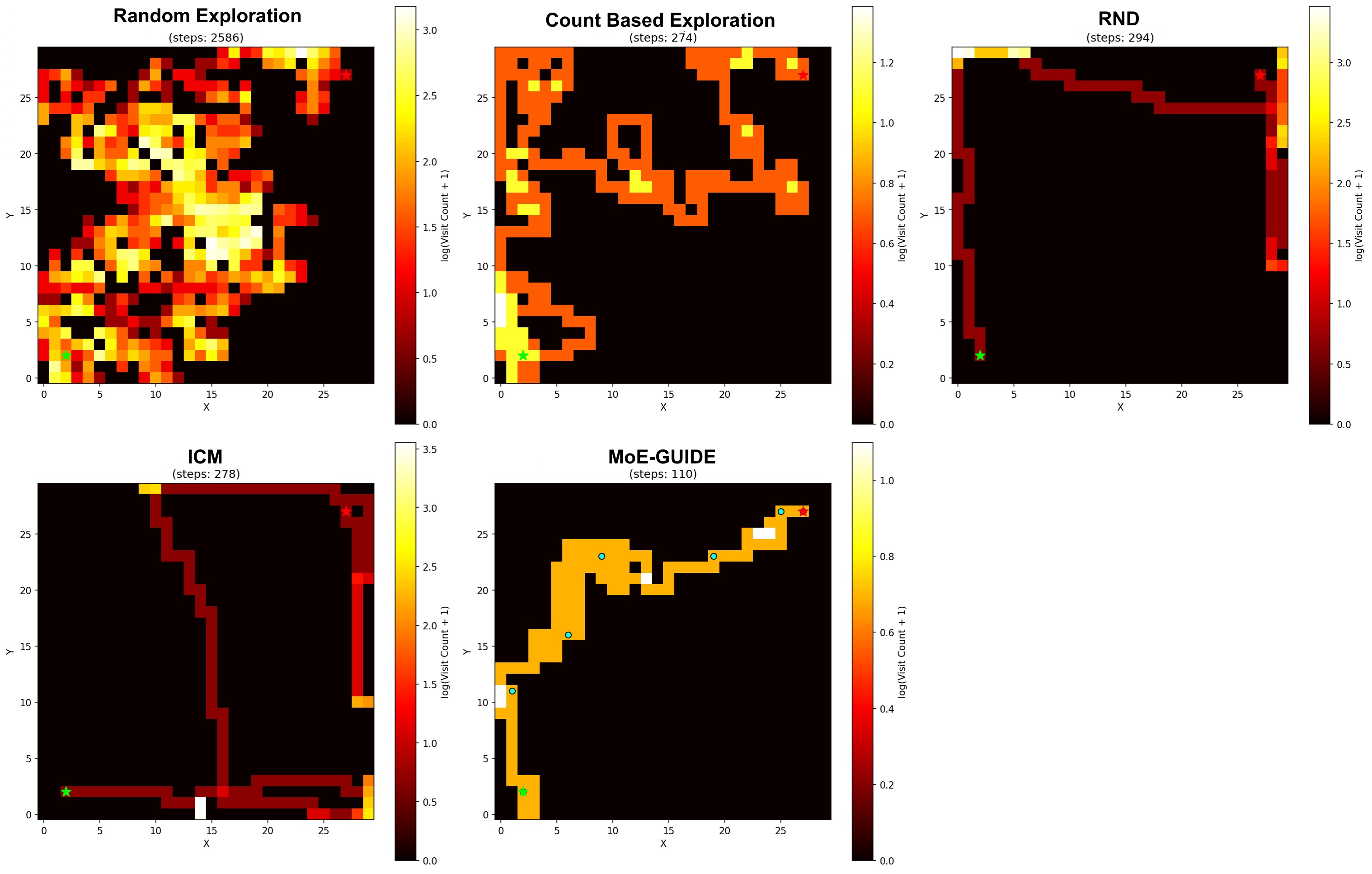}
    \caption{
        Exploration patterns in a gridworld with random walls for different intrinsic motivation methods: random, count-based, ICM, RND, and MoE-GUIDE. The agent always chooses among actions with the highest intrinsic reward, illustrating the characteristic exploration behavior of each method. For MoE-GUIDE, intrinsic rewards are only provided once per state to prevent the agent from getting stuck. The green star is the start state, the red star is the goal, and blue dots indicate demonstration states.
    }
    \label{fig:gridworld_methods}
\end{figure}

Figure~\ref{fig:gridworld_methods} visualizes the exploration patterns produced by different intrinsic motivation methods: random, count-based, ICM, RND, and MoE-GUIDE. This comparison highlights the distinct behaviors and exploration strategies induced by each method.

\begin{figure}[h!]
    \centering
    \includegraphics[width=1\textwidth]{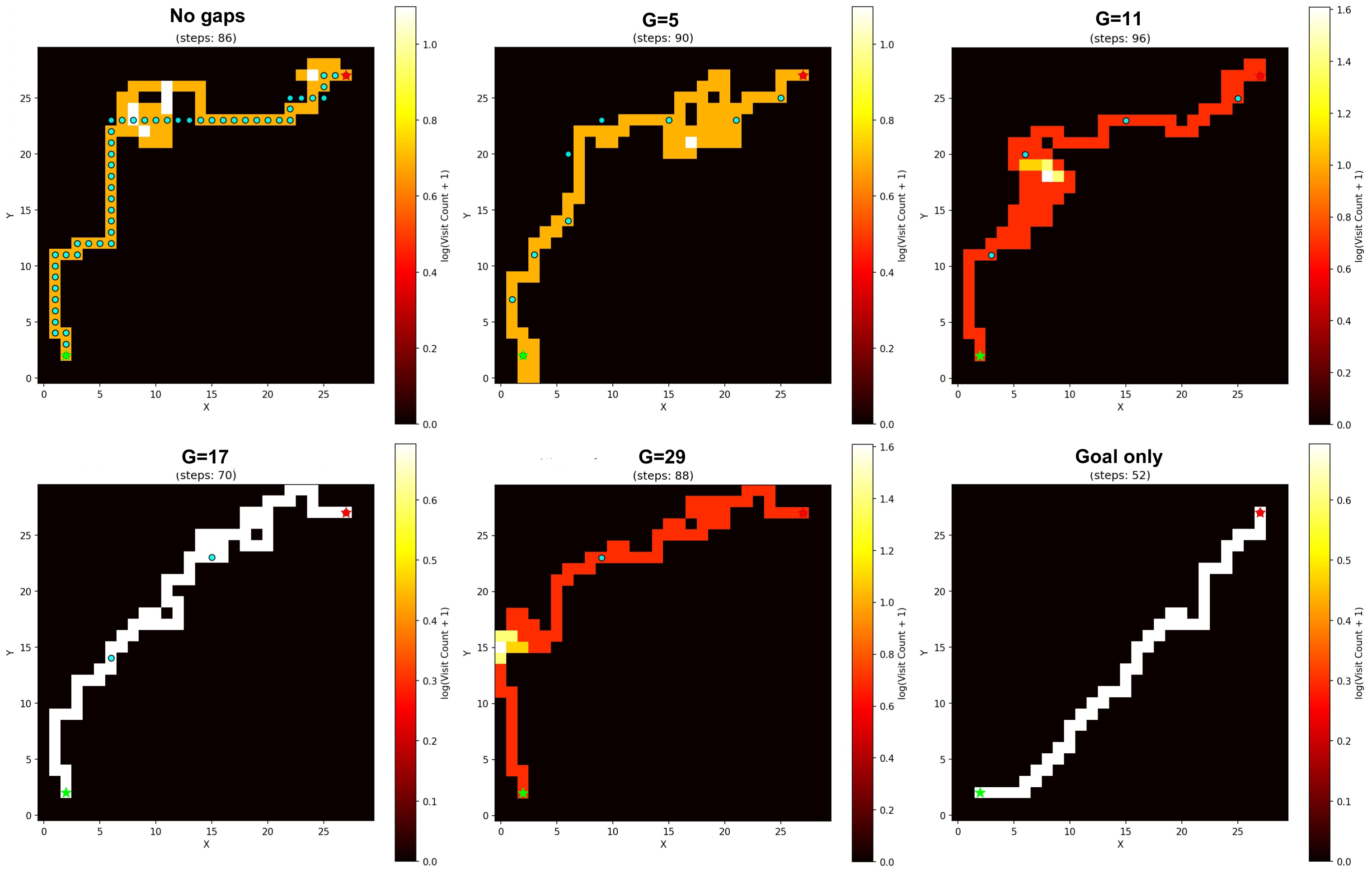}
    \caption{
        Effect of gaps in demonstration data on MoE-GUIDE’s exploration in gridworld. Each subplot corresponds to a different gap size $G$, where $G=\text{NUMBER}$ indicates the number of samples skipped between demonstration points. The agent clearly attempts to explore regions where demonstration states are present, even as the demonstration data becomes increasingly sparse. The green star is the start state, the red star is the goal, and the blue dots indicate demonstration states.
    }
    \label{fig:gridworld_gaps}
\end{figure}

Figure~\ref{fig:gridworld_gaps} compares MoE-GUIDE’s behavior under different amounts of missing data in the demonstrations. Each subplot corresponds to a different gap size $G$, where $G=\text{NUMBER}$ indicates the number of samples skipped between demonstration points. It is clearly visible that the agent attempts to explore regions where demonstration states are present, even when the demonstration data is sparse due to gaps.

\section{Tables of final mean results}
This section provides tables summarizing the final mean rewards and standard deviations for different experimental settings and hyperparameters. The results are presented to facilitate comparison between methods and configurations.

\begin{table}[h!]
\centering
\caption{Final mean rewards $\pm$ standard deviation for each method in the perfect expert experiment.}
\scriptsize
\begin{tabularx}{\textwidth}{lXXXXX}
\toprule
\textbf{Method} & \textbf{Swimmer} & \textbf{Hopper} & \textbf{Walker2d} & \textbf{HalfCheetah} & \textbf{Ant} \\
\midrule
ER+pretraining & 86.59 $\pm$ 15.25 & 2556.55 $\pm$ 551.80 & 3746.44 $\pm$ 1953.40 & 2112.67 $\pm$ 2250.98 & 4815.93 $\pm$ 324.09 \\
IR+pretraining & 321.32 $\pm$ 3.26 & 2994.18 $\pm$ 947.20 & 4841.49 $\pm$ 55.06 & -325.95 $\pm$ 147.75 & 4611.79 $\pm$ 140.51 \\
ER-only & 100.66 $\pm$ 38.36 & 2946.26 $\pm$ 672.87 & 4201.02 $\pm$ 646.91 & 11216.96 $\pm$ 508.09 & 3603.56 $\pm$ 1704.07 \\
MoE-GUIDE & 329.50 $\pm$ 1.70 & 3642.78 $\pm$ 196.37 & 4776.34 $\pm$ 168.72 & 9867.41 $\pm$ 907.54 & 5282.29 $\pm$ 222.13 \\
\bottomrule
\end{tabularx}
\end{table}

\begin{table}[h!]
\centering
\caption{Final mean rewards $\pm$ standard deviation for each method in the imperfect expert experiment.}
\scriptsize
\begin{tabularx}{\textwidth}{lXXX}
\toprule
\textbf{Method} & \textbf{Walker2d} & \textbf{HalfCheetah} & \textbf{Ant} \\
\midrule
ER-only          & 4200.66 $\pm$ 646.93   & 11225.46 $\pm$ 505.78   & 3603.56 $\pm$ 1704.66 \\
MoE-GUIDE        & 5046.87 $\pm$ 162.36   & 10829.20 $\pm$ 565.04   & 5020.49 $\pm$ 240.66 \\
IR+pretraining   & 4557.89 $\pm$ 120.08   & 5274.66 $\pm$ 3884.78   & 3865.11 $\pm$ 279.42 \\
ER+pretraining   & 4103.49 $\pm$ 1509.36  & 7564.65 $\pm$ 1229.63   & 4595.19 $\pm$ 513.56 \\
\bottomrule
\end{tabularx}
\end{table}

\begin{table}[h!]
\centering
\caption{Final mean rewards $\pm$ standard deviation for different decay rates.}
\scriptsize
\begin{tabularx}{\textwidth}{lXXX}
\toprule
\textbf{Method / Decay} & \textbf{Walker2d} & \textbf{HalfCheetah} & \textbf{Ant} \\
\midrule
0.999995    &    ---          & 10829.20 $\pm$ 565.04   & 4651.73 $\pm$ 424.67 \\
0.999996    &    ---         & 9377.68 $\pm$ 1383.32   & 4803.90 $\pm$ 375.54 \\
0.999997    & 4188.21 $\pm$ 1669.04  & ---                      & 4935.50 $\pm$ 349.51 \\
0.999998    & 4439.23 $\pm$ 1704.39  & ---                      & 4877.52 $\pm$ 383.50 \\
0.999999    & 5047.04 $\pm$ 162.74   & ---                      & 4951.07 $\pm$ 121.43 \\
ER-only     & 4200.63 $\pm$ 646.91   & 11223.27 $\pm$ 508.42 & 3603.59 $\pm$ 1704.69 \\
\bottomrule
\end{tabularx}
\end{table}

\begin{table}[h!]
\centering
\caption{Final mean rewards $\pm$ standard deviation for different $L_\mathrm{min}$ values.}
\scriptsize
\begin{tabularx}{\textwidth}{lXX}
\toprule
\textbf{$L_\mathrm{min}$} & \textbf{Extrinsic Reward} & \textbf{Intrinsic Reward} \\
\midrule
0.006 & 4139.03 $\pm$ 93.31   & 919.57 $\pm$ 13.74  \\
0.008 & 3812.84 $\pm$ 294.02  & 923.57 $\pm$ 73.03  \\
0.01  & 3372.02 $\pm$ 869.04  & 892.51 $\pm$ 158.13 \\
0.03  & -1812.56 $\pm$ 933.01 & 916.76 $\pm$ 114.37 \\
\bottomrule
\end{tabularx}
\end{table}

\begin{table}[h!]
\centering
\caption{Final mean rewards $\pm$ standard deviation for different gap sizes and number of demonstrations.}
\scriptsize
\begin{tabularx}{\textwidth}{lXX}
\toprule
\textbf{Setting} & \textbf{Intrinsic Reward + pretraining} & \textbf{MoE-GUIDE} \\
\midrule
l1\_g9    & ---              & 4199.43 $\pm$ 338.67 \\
l1\_g19   & 3010.78 $\pm$ 34.19    & 4271.73 $\pm$ 615.61 \\
l10\_g4   & 4611.81 $\pm$ 140.51   & 5282.29 $\pm$ 222.68 \\
l10\_g9   & 3871.64 $\pm$ 203.34   & ---                       \\
l10\_g14  & ---                       & 5208.85 $\pm$ 212.66 \\
l10\_g24  & 4177.66 $\pm$ 191.29   & 5050.68 $\pm$ 225.26 \\
ER-only   &                        & 3604.23 $\pm$ 1704.66 \\
\bottomrule
\end{tabularx}
\end{table}

\begin{table}[h!]
\centering
\caption{Final mean rewards $\pm$ standard deviations for each method in the sparse environment.}
\label{tab:sparse_env_rewards}
\begin{tabular}{lccc}
\toprule
\textbf{Method}     & \textbf{Walker2d}         & \textbf{HalfCheetah}      & \textbf{Ant} \\
\midrule
ER-only             & $23.50 \pm 11.85$         & $89.26 \pm 29.96$         & $4.33 \pm 8.47$ \\
ER+pretraining      & $23.04 \pm 16.13$         & $13.04 \pm 8.31$          & $76.39 \pm 157.01$ \\
IR+pretraining      & $48.83 \pm 7.77$          & $263.42 \pm 147.87$       & $530.58 \pm 35.52$ \\
MoE-GUIDE           & $34.95 \pm 48.01$         & $127.90 \pm 174.62$       & $526.62 \pm 5.14$ \\
ER+RND              & $14.16 \pm 28.72$         & $52.79 \pm 24.78$         & $-0.23 \pm 0.43$ \\
ER+ICM              & $-0.22 \pm 0.10$          & $56.64 \pm 67.66$         & $-0.12 \pm 0.09$ \\
\bottomrule
\end{tabular}
\end{table}

\end{document}